%% file: root.tex
\documentclass[letterpaper]{IEEEtran}
\usepackage{multicol}
\usepackage{cite}
\usepackage[bookmarks=true]{hyperref}
\usepackage{graphics} 
\usepackage{epsfig}
\usepackage{amsmath}
\usepackage{amssymb} 
\usepackage{mathrsfs}
\usepackage{amsfonts}
\usepackage{svg}
\usepackage{wrapfig}
\usepackage{graphicx}
\usepackage{float}
\usepackage{ctable}
\usepackage{cuted}
\usepackage{colortbl}
\usepackage{multirow}
\usepackage{siunitx}
\usepackage{threeparttable}
\usepackage{pifont}
\usepackage{hyperref}


\IEEEoverridecommandlockouts             
\title{\LARGE \bf
DexSinGrasp: Learning a Unified Policy for Dexterous Object Singulation and Grasping in Densely Cluttered Environments}

\newcommand{\TODO}[1][]{\textcolor{red}{\bf{TODO}}}

\renewcommand{\arraystretch}{1.5}

\definecolor{myblue}{rgb}{0.85, 0.92, 0.98}
\author{Lixin Xu$^{1*}$, Zixuan Liu$^{1*}$, Zhewei Gui$^{1}$, Jingxiang Guo$^{1}$, 
\\Zeyu Jiang$^{1}$, Tongzhou Zhang$^{1}$, Zhixuan Xu$^{1}$,  Chongkai Gao$^{1}$, Lin Shao$^{1\dagger}$
\thanks{* denotes equal contribution}
\thanks{$\dagger$ denotes the corresponding author}
\thanks{$^{1}$Lixin Xu, Zixuan Liu, Zhewei Gui, Jingxiang Guo, Zeyu Jiang, Tongzhou Zhang, Zhixuan Xu, Chongkai Gao, Lin Shao are with the School of Computing, National University of Singapore. \texttt{davidxulixin@gmail.com, zixuanliu@u.nus.edu, linshao@nus.edu.sg}
}}

\begin{document}

\maketitle

\thispagestyle{empty}
\pagestyle{empty}


\begin{abstract}

\input{tex/0-abs}
\end{abstract}

\section{Introduction} \label{sec: intro}
\input{tex/1-intro}
\section{Related Work}
\input{tex/2-related}
\section{Method}
\label{sec: method}
\input{tex/3-method}
\section{Experiment}

\input{tex/4-exp}
\section{Conclusion}
\input{tex/5-conclusion}

\begingroup
\renewcommand{\baselinestretch}{0.93}
\footnotesize
\setlength{\itemsep}{0pt}
{\tiny
\bibliographystyle{IEEEtran}
\bibliography{ref}
}
\endgroup

\end{document}

%% file: tex/0-abs.tex
Grasping objects in cluttered environments remains a fundamental yet challenging problem in robotic manipulation. While prior works have explored learning-based synergies between pushing and grasping for two-fingered grippers, few have leveraged the high degrees of freedom (DoF) in dexterous hands to perform efficient singulation for grasping in cluttered settings. In this work, we introduce \emph{DexSinGrasp}, a unified policy for dexterous object singulation and grasping. DexSinGrasp enables high-dexterity object singulation to facilitate grasping, significantly improving efficiency and effectiveness in cluttered environments. We incorporate clutter arrangement curriculum learning to enhance success rates and generalization across diverse clutter conditions, while policy distillation enables a deployable vision-based grasping strategy. To evaluate our approach, we introduce a set of cluttered grasping tasks with varying object arrangements and occlusion levels. Experimental results show that our method outperforms baselines in both efficiency and grasping success rate, particularly in dense clutter. Codes, appendix, and videos are available on our website~
\href{https://nus-lins-lab.github.io/dexsingweb/}{https://nus-lins-lab.github.io/dexsingweb/}.

%% file: tex/1-intro.tex
Dexterous grasping of target objects in cluttered environments is crucial for various applications, from production lines \cite{mohammed2022review} to assembly processes \cite{catgrasp,robograspclutter} and beyond. While dexterous hands offer high degrees of freedom (DoF) and substantial potential for complex manipulation tasks \cite{christenDGraspPhysicallyPlausible2022,zhangGraspXLGeneratingGrasping2024,manifm, dro,wanUniDexGraspImprovingDexterous2023,wangUniGraspTransformerSimplifiedPolicy2024}, effectively leveraging their capabilities for grasping in cluttered settings remains a challenging problem. The importance of dexterous grasping in cluttered environments lies in its ability to singulate surroundings efficiently, where the flexibility of fingers can be leveraged to create sufficient space for a successful grasp. Recent dexterous grasping approaches \cite{graspanythig,ddgc} focus primarily on grasping target objects in scenarios without the need to rearrange surrounding objects. These methods are typically designed for loosely cluttered environments, where they extract scene information from segmented point clouds \cite{graspanythig} or segmented images \cite{ddgc} to identify suitable grasping positions. However, due to the lack of explicit singulation training, these approaches struggle in denser clutter, where avoiding interaction with surrounding objects is insufficient to ensure grasp success.

One approach to handling densely cluttered environments is to \emph{singulate}---spatially separate---the target object from surrounding objects. Researchers have explored frameworks to learn the synergies between pushing and grasping \cite{zengLearningSynergiesPushing2018,xuEfficientLearningGoalOriented2021,slpushgrasp} for two-fingered grippers. These methods emphasize that in densely cluttered scenes where the target object is initially ungraspable, singulation must first be performed before grasping \cite{retriobjvis,liuGEGraspEfficientTargetOriented2022,zhangINVIGORATEInteractiveVisual2024,adagrasp, pixelation, grasp_transfer, grasp_wild,volgraspnet,splitdql}. Naturally, this leads to a pushing-and-grasping process, where Zeng et al. \cite{zengLearningSynergiesPushing2018} proposed a method using two networks to coordinate pushing and grasping. However, their approach relies solely on a grasp success reward to guide coordination, leading to low sample efficiency during training. Other methods introduced adversarial training to enhance synergy learning efficiency\cite{xuEfficientLearningGoalOriented2021,slpushgrasp}. Despite these advancements, the mechanical constraints of grippers require the target object to be fully isolated from surrounding clutter, making singulation inefficient. In contrast, dexterous hands perform singulation using only their fingers, minimizing movement of the end-effector (i.e., the palm) and providing a more flexible and efficient execution in cluttered settings.

\begin{figure}[t]
    \centering    \includegraphics[width=0.98\linewidth]{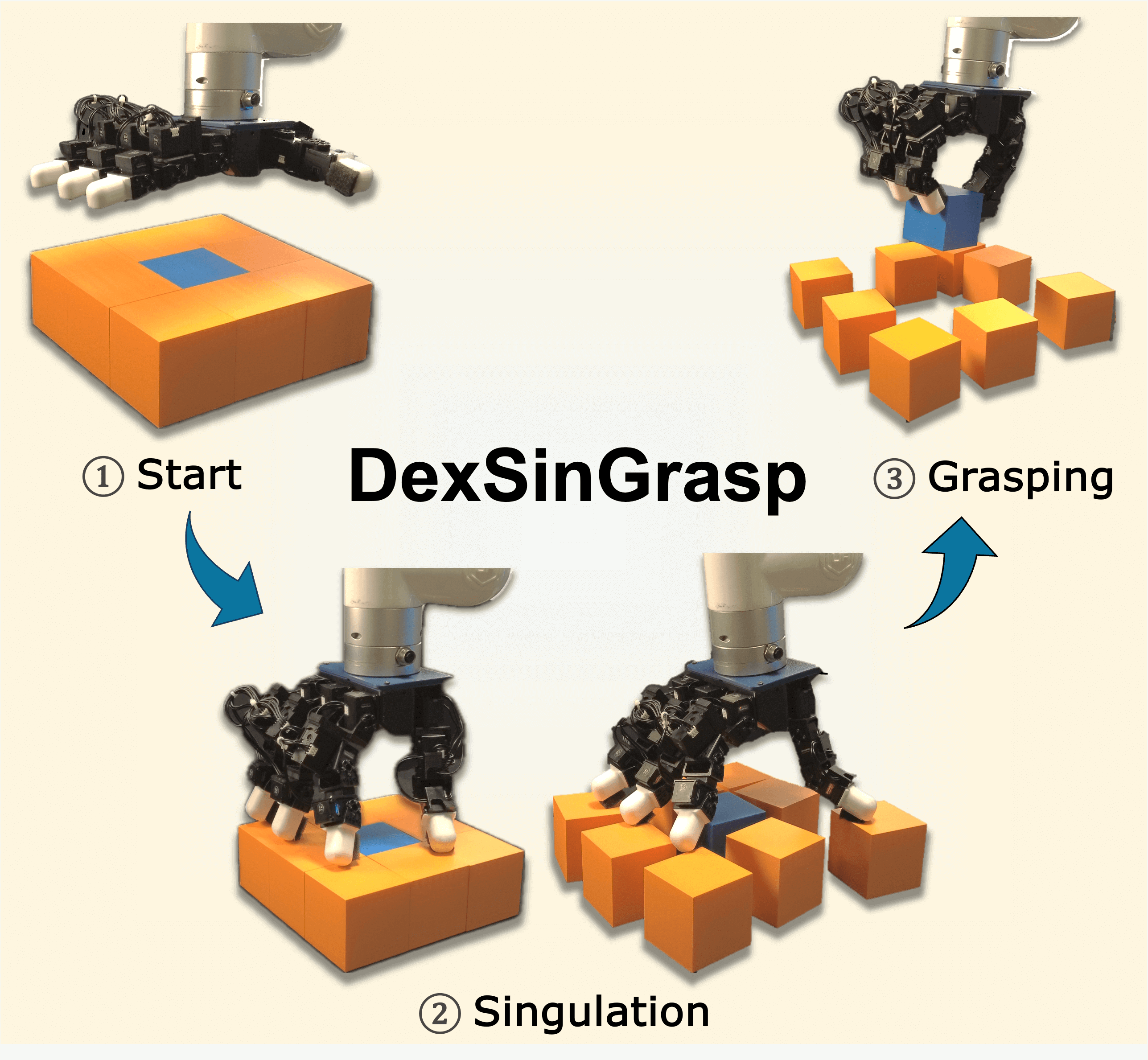}
    \vspace{-2mm}
    \caption{We propose DexSinGrasp to learn a unified policy for dexterous object
singulation and grasping in densely cluttered environments}
    \label{fig: teaser}
    \vspace{-6mm}
\end{figure}

\begin{figure*}[ht!]
    \centering
    \includegraphics[width=0.9\textwidth]{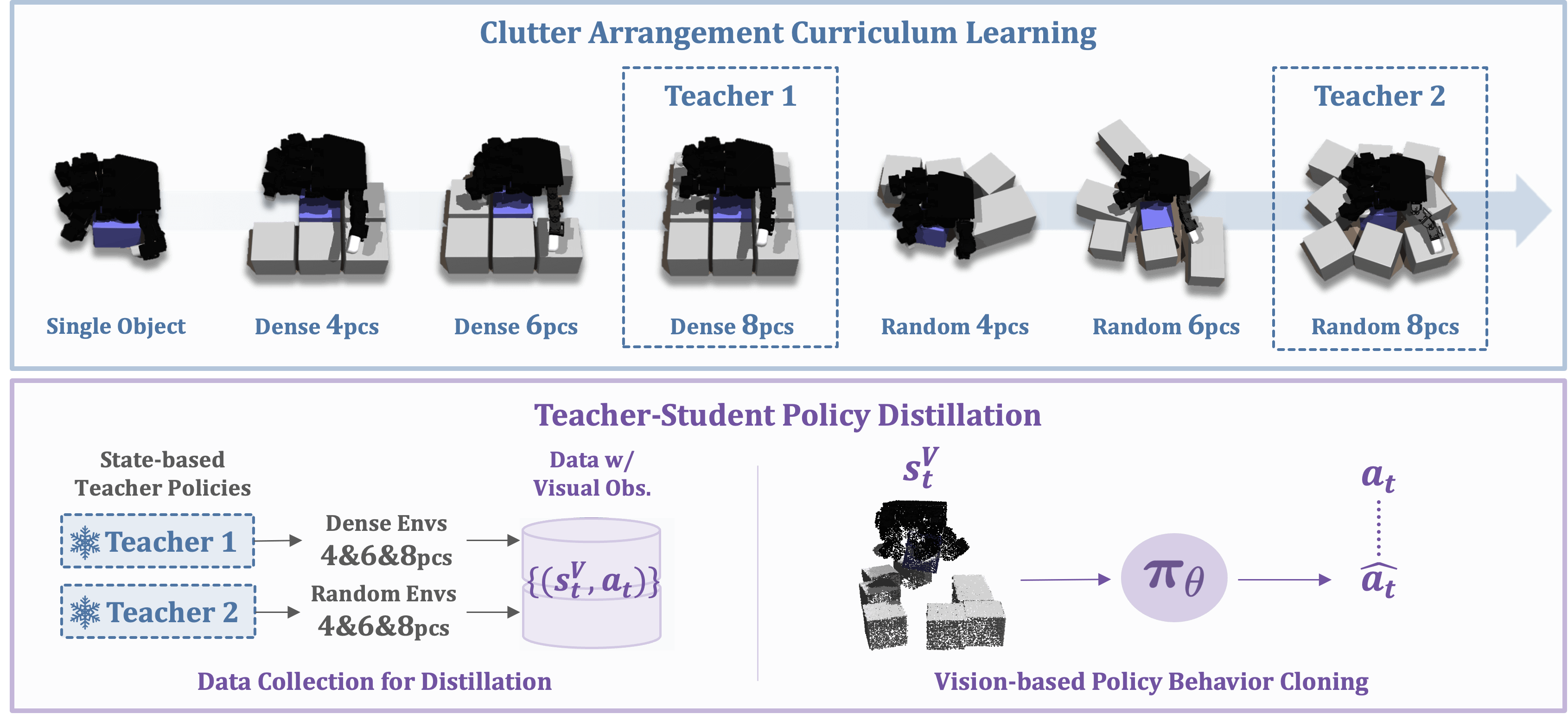}
    \vspace{-2mm}
    \caption{Framework of DexSinGrasp. 
   Firstly, we adopt clutter arrangement curriculum learning to progressively improve the performance of our teacher policy to address the challenge of training from scratch in dense or random clutter arrangements, and acquire two teacher policies for dense and random arrangement tasks, respectively.
   We then collect data with visual observation from these two teachers and finally train a vision-based student policy via behavior cloning, which better facilitates real-world deployment.
    }
    \label{fig: framework}
    \vspace{-6mm}
\end{figure*}
\label{Sec: prob_state}

Learning singulation for grasping with dexterous hands offers a promising approach to handling objects in cluttered environments. However, the high degrees of freedom (DoF) of dexterous hands and the complexity of cluttered scenes make this synergy challenging to learn. One approach to addressing this challenge is task decomposition. Jiang et al. \cite{dexsing} propose a three-stage framework where separate policies are trained for isolating, grasping, and lifting. While this decomposition simplifies learning by breaking the problem into manageable sub-tasks, it also limits the synergy between singulation and grasping. In particular, minimal singulation may often suffice for dexterous grasping, as illustrated in Fig. \ref{fig: teaser}.
Alternatively, curriculum learning \cite{Bengio2009CurriculumL} has proven effective in tackling complex tasks and has already been successfully applied to dexterous grasping policies \cite{xuUniDexGraspUniversalRobotic2023,wanUniDexGraspImprovingDexterous2023}. More recently, Bai et al.~\cite{bai2025retrieval} utilize a dexterous hand to expose the target object from clutter for grasping, but their reward design primarily encourages palm movement during singulation. In contrast, we argue that finger dexterity should be leveraged throughout the singulation to improve efficiency. For instance, in a top-down grasp, the palm can remain above the target while the fingers actively separate surrounding objects until a feasible grasp is possible. Therefore, our work builds on this success by exploring curriculum learning to develop a unified policy that integrates both singulation and grasping with finger dexterity.

In this work, the framework enables a dexterous hand to efficiently retrieve the target object from tightly cluttered environments, as illustrated in Fig. \ref{fig: teaser}. We develop a reinforcement learning (RL) framework to seamlessly integrate object singulation and grasping through a unified reward design. Due to the challenges of directly solving grasping tasks in densely cluttered environments, our method leverages clutter arrangement curriculum learning to progressively enhance the performance of the teacher policy in generated cluttered environments with increasing complexity in object quantity, types, and arrangements. Furthermore, through teacher-student policy distillation, we obtain a vision-based student policy that generalizes across diverse cluttered environments and can be deployed on a real-world robot.

\begin{itemize}
    \item We develop a unified RL policy for dexterous object singulation and grasping, enabling dexterous hands to effectively and efficiently retrieve objects in tightly cluttered environments.
    \item We incorporate clutter arrangement curriculum learning to improve policy performance across various cluttered scenes and employ policy distillation to obtain a vision-based policy suitable for real-world deployment.
    \item We design a set of cluttered grasping tasks and experiments with varying difficulty levels and conduct extensive experiments to demonstrate the effectiveness and efficiency of our proposed DexSinGrasp, and to verify the significance of finger dexterity.
\end{itemize}

%% file: tex/2-related.tex
\textbf{Dexterous Grasping Policy.} Dexterous grasping with high degrees of freedom (DoF) presents significant learning challenges. To address these challenges, some approaches decompose grasping into hierarchical levels and use large grasping dataset to train high-level grasp proposal module~\cite{christenDGraspPhysicallyPlausible2022,zhangGraspXLGeneratingGrasping2024,manifm,dro} in a supervised setting, followed by motion planning or low-level policies. Modern approaches on dexterous grasping benifit from RL and massive parallel simulations for dynamic hand–object interactions. Building on this foundation, others focus on obtaining deployable vision-based policies for real-world applications through policy distillation~\cite{wanUniDexGraspImprovingDexterous2023,wangUniGraspTransformerSimplifiedPolicy2024}. 
Researchers have also explored population-based training for reward-scale tuning, which can significantly reduce the
cost of manual hyperparameter tuning \cite{dexpbt}.
Recently, dexterous grasping in cluttered environments has gained increasing attention. Through encouraging visual exposure of the target object \cite{graspanythig}, and extracting scene information from segmented point clouds \cite{graspanythig} or segmented images \cite{ddgc}, grasping in relatively loose clutter conditions is successfully achieved. However, rather than solely focusing on grasping, our approach learns to perform object singulation for grasping, leading to a significantly higher success rate, especially in tightly cluttered environments. 

\textbf{Grasping in Object Clutters.} Prior studies on robotic grasping in cluttered environments with grippers have explored various approaches to address the challenges such as occlusions \cite{retriobjvis,liuGEGraspEfficientTargetOriented2022,zhangINVIGORATEInteractiveVisual2024} and limited grasping space \cite{adagrasp, pixelation, grasp_transfer, grasp_wild,volgraspnet, nonprehensile}. A common requirement in these scenarios is decluttering or singulation \cite{zengLearningSynergiesPushing2018,splitdql,slpushgrasp,huangDIPNDeepInteraction2021,pregraspretritar,nonprerearang,iim,kiatosLearningPushGraspingDense2022,rlclutter,robpnp,xuEfficientLearningGoalOriented2021,actvistac}, which isolates the target object to facilitate successful grasping. Zeng et al.~\cite{zengLearningSynergiesPushing2018} utilized two separate networks to learn the synergy between pushing and grasping with a parallel gripper, inspiring subsequent studies on the adversarial training of these two networks \cite{xuEfficientLearningGoalOriented2021,slpushgrasp}. Chen et al.~\cite{chen2023synthesizing} optimize contact and object trajectories and solve hand motion via inverse kinematics in reverse order, forming an open-loop control strategy, while our learning-based approach achieves adaptive closed-loop control, making it more effective in dynamic and contact-rich tasks. Due to large space needed for palm-driven pushing and limited efficiency of parallel grippers, SOPE \cite{dexsing} uses fingers of a dexterous hand to isolate the target object as one of the pre-defined stages. However, our method uses clutter arrangement curriculum learning to obtain a synergistic dexterous singulation that greatly improves grasping efficiency. Recent work \cite{bai2025retrieval} has explored using dexterous hands to expose target objects from clutter for grasping, but primarily relies on palm-driven motion during singulation. In contrast, we highlight the importance of using finger dexterity throughout the singulation to improve efficiency through our unified reward design.

\textbf{Curriculum Learning.} 
Curriculum learning has been proven to improve both task success rates and generalization \cite{Bengio2009CurriculumL}. In challenging dexterous grasping tasks, it has been used to design training curricula based on object geometry \cite{xuUniDexGraspUniversalRobotic2023,wanUniDexGraspImprovingDexterous2023}, significantly enhancing grasping generalization. In contrast, we use clutter arrangement curriculum learning to start from single object grasping to structured clutter distributions, thus reducing the difficulty of direct learning in fully compact or random clutter and leading to improved success rates and better generalization of the teacher policy.

%% file: tex/3-method.tex
\textbf{Problem Formulation} We first define a \emph{Singulation-Necessary Clutter} (SNC) as a clutter configuration in which the target object is initially ungraspable and singulation is required before any feasible grasp can be executed. 

To construct SNC scenes in a controlled and reproducible way, we impose the following three sufficient conditions: (1) The target object is fully enclosed by non-target objects and located at the geometric center of the clutter; (2) The horizontal distance between the target and each surrounding object is smaller than the fingertip size; (3) The target and surrounding objects have identical height.

These constraints ensure that direct grasping is physically infeasible, and that singulation is required to create graspable space. Based on the conditions, we consider a 2D cluttered scene on a horizontal tabletop, where the target object lies in the same plane as surrounding distractor objects, with no vertical occlusion. The densely cluttered setting mentioned in the Sec.~\ref{sec: intro} serves as a concrete example of SNC. We further study two representative types of SNC:

\textbf{Dense Arrangement.} Objects are tightly packed with minimal spatial variability due to strong geometric constraints.

\textbf{Random Arrangement.} Objects are more loosely placed with slightly increased inter-object distances, resulting in more diverse spatial configurations.

Throughout training, we use cuboid-shaped objects as the standard shape for training. This choice facilitates the design of our curriculum learning scheme by providing consistent and controllable geometric structure. Moreover, we empirically find that policies trained on cuboids exhibit robust generalization to unseen shapes during later evaluations.

We then formulate dexterous grasping and object singulation as a RL task. Specifically, we consider a tabletop scene with the target object $b^\text{target}$ and $n$ surrounded objects $\left\{b_i\right\}_{i=1}^n$. To overcome this challenge, we train a policy to singulate the target object from its surroundings, thereby creating sufficient space for inserting fingers during dexterous grasping. Once singulation and grasping is complete, the target object is lifted to a goal position above the table. As a solution, we propose DexSinGrasp, a unified policy that jointly optimizes dexterous object singulation and grasping using RL. 

\paragraph{Observation Space}
The observation space of this singulation and grasping is defined as 
\begin{equation}
    s_t \triangleq  \left[ s_t^R, a_{t-1},s_t^O,d_t^{HO},T_t, d_t^S \right] \in \mathbb{R}^{168},
\end{equation}
where the proprioceptive robot state $s_t^R\in\mathbb{R}^{72}$ includes the wrist pose as well as joint positions, velocities, and forces for each finger and wrist dummy joints (All joint-related dimensions hereafter are based on the Leap Hand \cite{shaw2023leap} and omitted for brevity); the action $a_{t-1}$ at the previous time step will be discussed later; the object state $s_t^O\in\mathbb{R}^{16}$ consists of the object's position and quaternion, linear and angular velocity, and object-hand position difference; the hand-object distances $d_t^{HO}\in\mathbb{R}^{21}$ present the minimum distances between each hand links and points on the object; the time encoding $T_t\in\mathbb{R}^{29}$ encodes the current time along with a sine-cosine time embedding. The
singulation distance $d_t^S \in\mathbb{R}^{8}$ presents the distances between the target object and surrounding objects, indicating the level of enclosure within the clutter. If the number of surrounding objects satisfies $n<8$, the corresponding dimensions are padded with 0.

\paragraph{Action Space}
The action space $a_t \triangleq \left[ a_t^P, a_t^F\right]\in\mathbb{R}^{22}$ includes palm delta pose $a_t^P\in\mathbb{R}^{6}$ and linearly smoothed finger joint positions $a_t^F:= \lambda a_t^F + (1-\lambda) a_{t-1}^F\in\mathbb{R}^{16}$ for each finger.

The reward function, as a crucial role of our unified policy, will be discussed in detail in Sec. \ref{Sec: Unifying}.

\textbf{Overview.} As illustrated in Fig. \ref{fig: framework}, we train a unified policy for dexterous object singulation and grasping through a structured learning framework, and adopt teacher-student policy distillation for real-world deployment. First, we introduce a unified reward design (Sec. \ref{Sec: Unifying}) that seamlessly integrates singulation and grasping into a single objective, enabling more efficient policy learning. Then, to improve learning efficiency in SNCs, we adopt Clutter Arrangement Curriculum Learning (Sec. \ref{Sec: Curriculum}) to progressively train state-based teacher policies. Finally, we employ Teacher-Student Policy Distillation (Sec. \ref{Sec: student}) to transfer knowledge from the teacher policies to a vision-based student policy, allowing deployment on a real robot by mapping high-dimensional visual observations to effective actions.

\begin{table*}[htbp!]
    \centering
    \renewcommand{\arraystretch}{1.2} 
    \caption{Reward-related terms}
    \label{tab: reward}
    \vspace{-2mm}
    \begin{tabular}{ccl}
    \toprule
        Term & Equation & Explanation \\
    \midrule
        $d^P$ & $\min_{i=1}^p\Vert p^\text{palm}-p_i^\text{target}\Vert_2$ & Minimized distance of palm to target object. \\
        $r^P$ & $-2.0\times d^P$ & Palm reward to encourage hand approaching the target object. \\
        $d^J$ & $\sum_{j=1}^m\min_{i=1}^p\Vert p_j^\text{link}-p_i^\text{target}\Vert_2$ & Minimized summed distance of hand links to target object. \\
        $r^J$ & $-d^J$ & Joint reward to encourage the finger to grasp the target object. \\
        $r^F$ & $-\sum_{j=1}^h\min_{i=1}^p\Vert p_j^\text{fingertip}-p_i^\text{target}\Vert_2$ & Fingertip reward to encourage fingertip to grasp the target object. \\
        $r^L$ & $0.2 + 0.6 \times a^{P_{t_{z}}}$ & Lifting reward to encourage lifting actions. \\
        $r^G$ & $0.9-2.5\times\Vert p^\text{goal}-p^\text{target}\Vert_2$ & Goal reward to encourage target object to approach the goal. \\
        $r^S$ & $ 50 \times\min_{i=1}^{n} \Vert p^\text{target}-p_i\Vert_2$ & Singulation reward for synergistic dexterous singulation.\\
        $r^B$ & $(1+10\times\Vert p^\text{goal}-p^\text{target}\Vert_2)^{-1}$ & Bonus term to target object to approach the goal. \\
    \bottomrule
    \end{tabular}
    \vspace{-2mm}
\end{table*}

\subsection{Unifying Dexterous Object Singulation and Grasping}\label{Sec: Unifying}
Training a policy for dexterous grasping from scratch in SNCs poses significant challenges due to the high-dimensional action space and complex contact interactions. A naive approach trains a policy with rewards only for successful grasps in SNCs, but this suffers from low sample efficiency due to the need to learn singulation and grasping simultaneously. Alternatively, a two-stage method trains separate policies for singulation and grasping, but it leads to longer execution times and suboptimal transitions, as it fails to leverage their synergy.

To address these issues, we propose a unified reward design that seamlessly integrates dexterous object singulation and grasping into a single learning objective. The piece-wise reward function is defined as
\begin{equation}
    r_t =
    \begin{cases} 
    r_t^P + r_t^J + r_t^S, &\text{if } d_t^P \geq 0.06 \text{ or } d_t^J \geq 0.2, \\
    \begin{aligned}
    r_t^P &+ r_t^J + r_t^F + r_t^L  \\ &+ r_t^G + r_t^S + r_t^B,
    
    \end{aligned} & \text{if } d_t^P < 0.06 \text{ and } d_t^J < 0.2,
    \end{cases}
\end{equation}

Detailed definitions of these rewards can be found in Tab.~\ref{tab: reward}, where the subscript $t$ is omitted for simplicity. The rewards are defined as follows: $r_t^P$ encourages the palm to approach the target object; $r_t^J$ and $r_t^F$ promote grasp formation through joint and fingertip contacts; $r_t^L$ encourages lifting once contact is established; $r_t^S$ drives separation of the target from obstacles; $r_t^G$ guides motion toward the goal position; and $r_t^B$ provides a bonus for successful singulation and grasping. $d_t^P$ and $d_t^J$ denote the minimum distances from the palm and hand links to the target, respectively. $\{p_i^{\text{target}}\}_{i=1}^{N_p}$ are the target surface points; $p^{\text{palm}}$, $\{p_j^{\text{link}}\}_{j=1}^m$, and $\{p_j^{\text{fingertip}}\}_{j=1}^h$ denote the palm, link, and fingertip positions; $p^{\text{goal}}$ and $p^{\text{target}}$ are the goal and current positions of the target object; $\{p_i\}_{i=1}^n$ are the obstacle positions; and $a^{P_{t_z}}$ represents palm translation along $+z$, corresponding to the lifting motion.

The reward function consists of two main components: an approach reward that encourages the hand to move toward the target object and a lifting reward that promotes object elevation after contact is established. To further facilitate singulation, the singulation reward $r_t^S$ is incorporated into both components, incentivizing the hand to separate the target object from surrounding obstacles. The transition between these two reward stages is achieved by a contact criterion specified by $d_t^P<0.06$ and $d_t^J<0.2$. These values are selected such that the palm and fingers are close enough for grasping. The reward transitions from approaching to grasping, ensuring a smooth progression toward a successful grasp. 

Despite this unified learning framework, training remains highly challenging due to the increasing complexity of clutters. To further improve learning efficiency and policy generalization, we introduce Clutter Arrangement Curriculum Learning, which progressively increases clutter complexity. 

\subsection{Clutter Arrangement Curriculum Learning}\label{Sec: Curriculum}
To ensure successful and efficient object singulation and grasping in SNCs, we begin by training the teacher policy with privileged information from the simulation and adopt clutter arrangement curriculum learning, allowing our teacher policy to progressively improve as object diversity and spatial complexity increase. Using Proximal Policy Optimization (PPO) \cite{ppo}, we optimize the policy to maximize the cumulative discounted reward \( E[\sum_{t=1}^{T}\gamma^{t-1}r_t] \), enabling effective RL.

\subsubsection{Clutter Generation}\label{Sec: env}
\begin{figure}[t!]
    \centering
    \includegraphics[width=\linewidth]{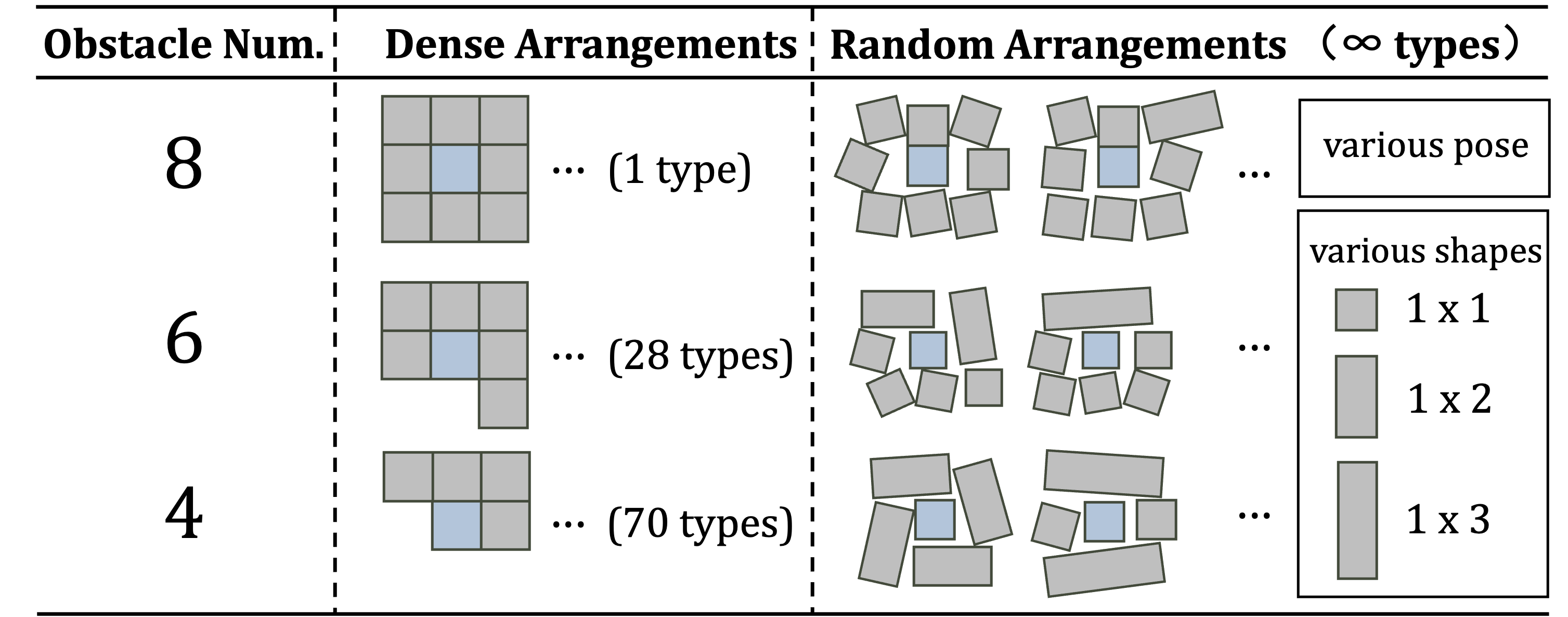}
    \vspace{-6mm}
    \caption{Dense and random arrangement settings. We introduce a cluttered environment generation module to create diverse object settings.}
    \label{fig: arrangements}
    \vspace{-4mm}
\end{figure}

To facilitate the training and testing of policies across various object configurations, we introduce a clutter generation module designed to create diverse object-based tasks.
Our clutters consist of cuboid-shaped objects with varying quantities (from 4 to 8) and shapes (1×1, 1×2, and 1×3 cuboids). The generation for two types of clutters are introduced below, as shown in Fig. \ref{fig: arrangements}.

\textbf{Dense Arrangements Generation.}
This type of task arranges different quantities of the surrounding 1×1 cuboids densely near the target object to create an extreme scenario that challenges the singulation and grasping policies under dense and narrow conditions.

\textbf{Random Arrangements Generation.}
This type of task arranges objects of different quantities and shapes randomly around the target object for grasping, mainly to test the generalization of the singulation and grasping policies.

For simplicity, we use D/R-$n$ to denote task setting with $n$ objects for dense (D) or random (R) arrangements, such as D-8. In our curriculum, only D-8 and R-8 are SNC tasks, while D-4/6 and R-4/6 are intermediate stages to aid learning and are not necessarily SNCs.

\subsubsection{Clutter Arrangement Curriculum Learning}\label{Sec: curriculum}
Since it is challenging to learn object singulation and grasping with dexterous hands in compact or diverse environments, such as the D-8 or R-8 tasks, we design clutter arrangement curriculum to gradually increase the object diversity and spatial complexity. We begin by training a grasping policy designed exclusively for Single-Object (SO) scenarios, where the objective is to grasp a single cuboid. Based on this initial policy, we continuously follow the curriculum and train on increasingly complex singulation and grasping tasks. Specifically, we first train on dense arrangements with D-4, then D-6, and finally D-8 tasks. We then use the expert obtained from the D-8 training as the starting point for training on random arrangements with R-4, R-6, and R-8 tasks. At the end of the training process, we extract the final policies from the D-8 training and the R-8 training as two teacher policies.

\subsection{Teacher-Student Policy Distillation}\label{Sec: student}

Since privileged observations—such as object states and singulation distances—are difficult to obtain in the real world, and some proprioceptive data, like finger-joint forces, are limited by hardware constraints, we learn a vision-based student policy to ensure feasible real-world deployment.

\subsubsection{Data Collection for Distillation}
The data collection phase involves preparing training data using two distinct teacher policies: the dense-clutter teacher policy for the D-4/6/8 tasks and the random-clutter teacher policy for the R-4/6/8 tasks, respectively. In total, 1000 episodes of observation and action data, along with scene point cloud, are prepared as $\{(s_t^V,a_t)\}$, where $s_t^V$ will be discussed later. The dataset is structured so that the D-4 and R-4 tasks each account for 10\% of the total, while the D-6, D-8, R-6, and R-8 tasks each contribute 20\%. This balanced distribution better captures SNC tasks and ensures a comprehensive representation of varying task complexities, which is critical for effective policy distillation and better generalizability.

\subsubsection{Vision-Based Policy Behavior Cloning}
Our vision-based student policy uses scene point cloud instead of object poses, which cannot be accurately obtained in such heavily occluded SNCs. We specifically use behavior cloning to train the student policy, with data collected from two teachers. The different visual observation $s_t^V$ for the vision-based student policy is defined as
\begin{equation}
    s_t^V\triangleq \left[ s_t^R, a_{t-1},s_t^{O'},d_t^{HO},T_t,v^t\right] \in \mathbb{R}^{275}
\end{equation}
where singulation distance $d_t^S$ is removed from $s_t$ in the state-based teacher policy, and object state $s_t^O\in\mathbb{R}^{16}$ is substituted with center position of scene point cloud $s_t^{O'}\in\mathbb{R}^{3}$. Moreover, the vision-based policy includes the visual features $v_t\in\mathbb{R}^{128}$ encoded from the scene point cloud using a pre-trained point cloud encoder from UniGraspTransformer \cite{wangUniGraspTransformerSimplifiedPolicy2024}, with the encoder weights frozen during the policy distillation.

%% file: tex/4-exp.tex
In this section, we conduct comprehensive experiments to evaluate our proposed method, DexSinGrasp, in both simulation and real-world tasks. Through these experiments, we aim to address the following key questions: (1) How effective and efficient is our method for grasping in SNCs? (2) How does our method generalize to different spatial arrangements? (3) How effective is our clutter arrangement curriculum learning? (4) How does our method generalize to different object geometries? (5) How does finger dexterity affect singulation and grasping performance in SNCs? (6) How does our method perform on real-world tasks?

\vspace{-2mm}
\subsection{Baselines}

To evaluate our approach, we compare our proposed approach with two baseline methods:

\textbf{GraspReward-Only Method.} In this baseline, pure dexterous grasping is conducted without singulation. This baseline is trained from scratch in a single target object environment with the singulation reward set to zero \cite{wanUniDexGraspImprovingDexterous2023,wangUniGraspTransformerSimplifiedPolicy2024}.

\textbf{Multi-Stage Method}
This baseline is a two-stage framework where separately trained singulation and grasping policies operate in sequence as adopted by SOPE \cite{dexsing}. We train the separate singulation policy without the grasping reward stage as mentioned in Sec.~\ref{Sec: Unifying}. The singulation stage is switched to the grasping stage when $\sum_{i=1}^n\Vert p^\text{target}-p_i\Vert_2/n>0.16$, where $n$ is the number of surrounding objects.

\vspace{-2mm}
\subsection{Evaluation Metrics}
To evaluate the performance of our approach, we introduce two key metrics that assess both effectiveness and efficiency in dexterous object singulation and grasping.

\textbf{Success Rate.} The proportion of trials where the target object reaches the predefined goal position above the table, defined as $\Vert p^\text{goal}-p^\text{target}\Vert_2 < 0.05$. Denoted as SR.

\textbf{Average Steps.} The average number of steps to singulate and grasp the target object, denoted as AS. Failed episodes are counted as 300 steps (the maximum per episode).

The best results in each table are highlighted in bold.  Parenthesized values show the difference from our method.

\vspace{-2mm}
\subsection{Implementation Details} 
We use Isaac Gym for clutter arrangement curriculum learning. For each D/R-$n$ task, we used 1000 simulated environments and trained the PPO policy network over 10K iterations with a learning rate of $3\times10^{-4}$. We then evaluated and selected the best-performing iteration as the policy for the next-stage clutter arrangement curriculum learning. The student policy is trained over 200 epochs with a batch size of 12 trajectories, each composed of 300 steps of recorded simulation data and a learning rate of $1\times10^{-4}$. In both simulation and real-world experiments, hand rotation is disabled, restricting the end-effector to translation and finger motion. This encourages dexterous finger use for singulation and grasping, prevents unrealistic floating-hand movements that may cause arm–table collisions on the xArm, and eases sim-to-real transfer. We use Isaac Gym’s default friction (1.0) for experiments without domain randomization, and maintain small inter-object gaps (1–1.5 cm) to enable horizontal contact forces and avoid self-locking.

\begin{table}[t!]
\centering
\caption{Evaluation on dense arrangements.}\label{tab:training_task}
\vspace{-2mm}
\resizebox{\linewidth}{!}{
\begin{tabular}{@{}cc|cccc|cccc@{}}
\toprule
\multicolumn{2}{c|}{\multirow{2}{*}{Method}} & \multicolumn{4}{c|}{SR(\%)$\uparrow$} & \multicolumn{4}{c}{AS$\downarrow$} \\ \cmidrule(l){3-10}  & & D-4        & D-6       & \multicolumn{1}{c|}{D-8}     & Avg. & D-4 & D-6 & \multicolumn{1}{c|}{D-8} & Avg. \\ \midrule
\multicolumn{1}{c|}{\multirow{3}{*}{State-Based}} & GraspReward-Only              &  66    &   40  &  \multicolumn{1}{c|}{10}   &   39  &   202   &  252   &   \multicolumn{1}{c|}{292}  &   249   \\
\multicolumn{1}{c|}{} & Multi-Stage    &   77   &  76  &  \multicolumn{1}{c|}{64}   &   72  &   199  &  210   &  \multicolumn{1}{c|}{235}   &   215   \\
\multicolumn{1}{c|}{} & Ours (Teacher)          &  \textbf{98}    &  \textbf{99}   &   \multicolumn{1}{c|}{\textbf{97}}  &  \textbf{98}    &  \textbf{100}   &   \textbf{115}  &    \multicolumn{1}{c|}{\textbf{138}} &     \textbf{118} \\ \midrule
\multicolumn{1}{c|}{Vision-Based} & Ours (Student)   &  90      &  92     &    \multicolumn{1}{c|}{84}  &  89   &  122    &   123  &   \multicolumn{1}{c|}{159}  &  135        \\ \bottomrule
\end{tabular}
}
\vspace{-2mm}
\end{table}

\begin{table}[t!]
\centering
\caption{Evaluation on random arrangements.}\label{tab: random}
\vspace{-2mm}
\resizebox{\linewidth}{!}{
\begin{tabular}{@{}cc|cccc|cccc@{}}
\toprule
\multicolumn{2}{c|}{\multirow{2}{*}{Method}} & \multicolumn{4}{c|}{SR(\%)$\uparrow$} & \multicolumn{4}{c}{AS$\downarrow$} \\ \cmidrule(l){3-10} 
                      &  & R-4  & R-6 & \multicolumn{1}{c|}{R-8} & Avg. & R-4 & R-6 & \multicolumn{1}{c|}{R-8} & Avg. \\ \midrule
\multicolumn{1}{c|}{\multirow{3}{*}{State-Based}} & GraspReward-Only              &   73  &  61  &   \multicolumn{1}{c|}{33}  & 56     &  179   &  207   &   \multicolumn{1}{c|}{261}  &    216  \\
\multicolumn{1}{c|}{} & Multi-Stage   &   88   &   72  &  \multicolumn{1}{c|}{78}   &  79    &  156   &  170   &  \multicolumn{1}{c|}{178}   &   168   \\
\multicolumn{1}{c|}{} & Ours (Teacher)          &   \textbf{97}   &  \textbf{96}   &  \multicolumn{1}{c|}{\textbf{94}}   &  \textbf{96}    &  \textbf{94}  &  \textbf{95}   &  \multicolumn{1}{c|}{\textbf{103}}   &  \textbf{97}   \\
\midrule \multicolumn{1}{c|}{Vision-Based} & Ours (Student)    &   91   &    86 &   \multicolumn{1}{c|}{88}  &   88   & 101    &   113  &  \multicolumn{1}{c|}{114}   &  109    \\ \bottomrule
\end{tabular}
}
\end{table}

\vspace{-2mm}
\subsection{Main Results and Analysis}

We tested the dense-clutter teacher policy and the student policy on D-4/6/8 tasks and compared their performance with all the baselines. 
All methods were evaluated for 100 times. 

Based on the results presented in Tab.~\ref{tab:training_task}, the multi-stage singulation policy demonstrates a higher success rate than the GraspReward-Only baseline, suggesting that the singulation stage plays a positive role in task performance. Our dense-clutter teacher policy achieves a significantly higher average success rate of 98\%, with a substantially lower AS compared to the Multi-Stage policy. While the student policy has a lower SR than the teacher, it still outperforms the baselines and maintains a comparable AS. Our results show that combining separate singulation and grasping policies increases action steps and reduces efficiency, whereas our unified policy balances effectiveness and efficiency for grasping in cluttered environments including SNCs, thus addressing Q1.

In response to Q2, we evaluate the random-clutter teacher policy and the student policy on R-4/6/8 tasks with different object arrangements to test the spatial generalization ability of our policy. In Tab.~\ref{tab: random}, we first observe that baselines perform better in random than in dense arrangements, where looser gaps allow direct grasps. Our teacher policy can achieve an average SR of 96\% across all tasks. While the student policy achieves a lower SR than the teacher policy, it still outperforms all the baseliness in both effectiveness and efficiency.

\begin{figure}[t!]
    \centering
\includegraphics[width=0.5\textwidth]{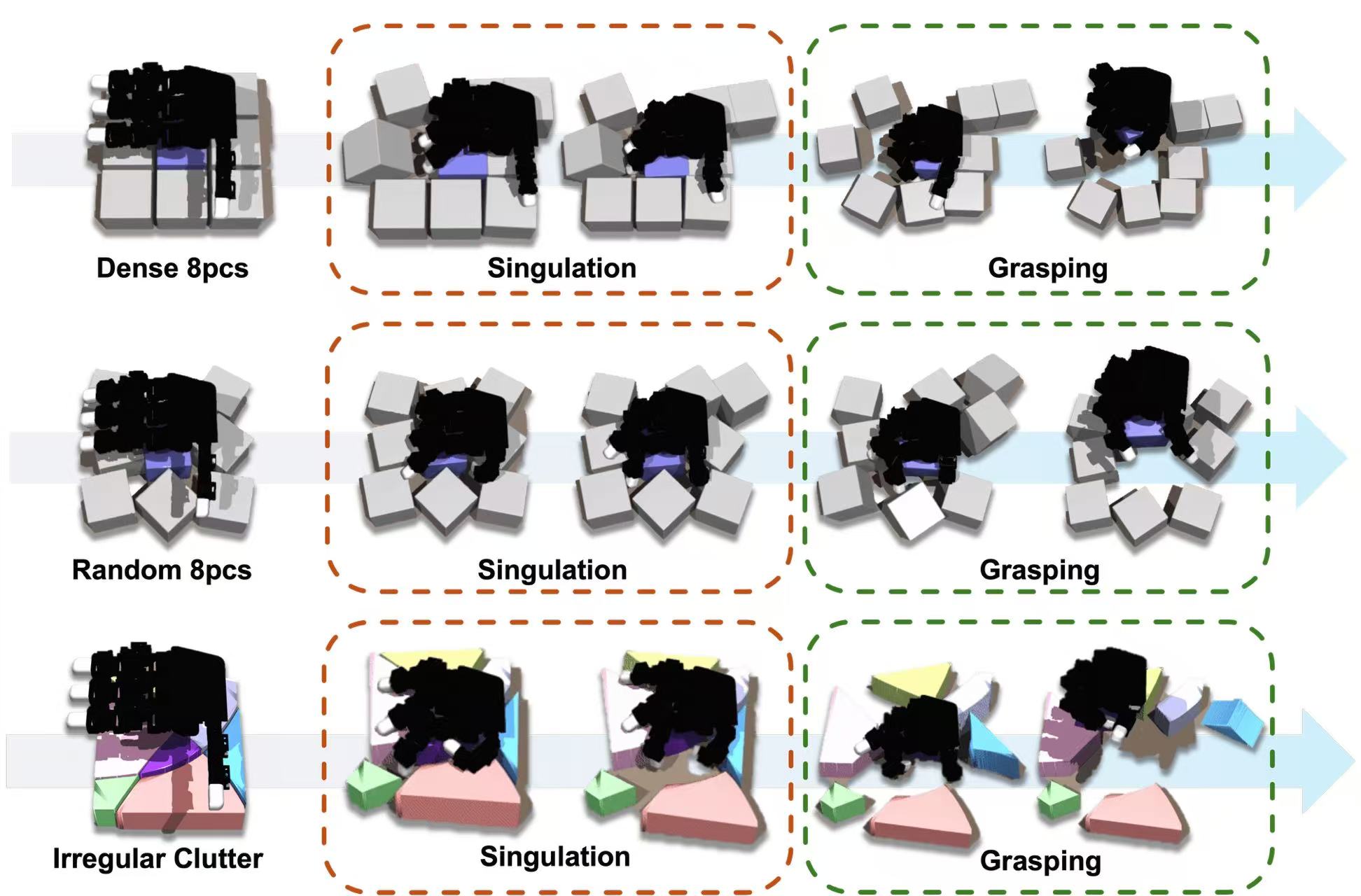}
    \vspace{-2mm}\caption{Qualitative results on object singulation and grasping in simulation.}
    \label{fig:quantitive}
    \vspace{-1mm}
\end{figure}

During training and testing, we found the policy learned several singulation patterns, including finger flickering and finger-palm vibration to displace and nudge surrounding objects, effectively singulating targets from clutters, as shown in Fig. \ref{fig:quantitive}. These patterns rely heavily on the finger dexterity, and a more detailed discussion is provided in Sec.~\ref{sec: dexterity}.

\begin{table}[t!]
\centering
\caption{Evaluation on different curriculums.}\label{tab: curriculum}
\vspace{-2mm}
\resizebox{\linewidth}{!}{
\begin{tabular}{@{}c|cccc|cccc@{}}
\toprule
\multirow{2}{*}{Curriculum} & \multicolumn{4}{c|}{SR(\%)$\uparrow$ on D-$n$ tasks} & \multicolumn{4}{c}{SR(\%)$\uparrow$ on R-$n$ tasks} \\ \cmidrule(l){2-9} 
                        & D-4  & D-6 & \multicolumn{1}{c|}{D-8} & Avg. & R-4 & R-6 & \multicolumn{1}{c|}{R-8} & Avg. \\ \midrule
Training from scratch         &   90 (8$\downarrow$)   &  75 (17$\downarrow$)   &  \multicolumn{1}{c|}{\textbf{97} (0)}   &  87 (9$\downarrow$)    & 74 (12$\downarrow$)   &  26 (70$\downarrow$)   &  \multicolumn{1}{c|}{0 (94$\downarrow$)}   &  33 (62$\downarrow$)   \\
Random-to-dense &   96 (2$\downarrow$)   &   87 (5$\downarrow$)  &  \multicolumn{1}{c|}{81 (16$\downarrow$)}   &  88 (8$\downarrow$)    &  \textbf{97} (1$\uparrow$)   &  92 (4$\downarrow$)   &  \multicolumn{1}{c|}{\textbf{97} (3$\uparrow$)}   &   \textbf{95} (0)   \\
Dense-to-random (Ours)     &   \textbf{98}   &    \textbf{92} &   \multicolumn{1}{c|}{\textbf{97}}  &   \textbf{96}   & 96    &   \textbf{96}  &  \multicolumn{1}{c|}{94}   &  \textbf{95}    \\ \bottomrule
\end{tabular}
}
\vspace{-2mm}
\end{table}

\vspace{-2mm}
\subsection{Clutter Arrangement Curriculum Learning Analysis}

The clutter arrangement curriculum learning process is designed to enhance success rates in increasingly complex scenes. 
We evaluate each policy trained on D/R-$n$ tasks under various curriculum directions—dense to random (SO, D-4, D-6, D-8, R-4, R-6, R-8), random to dense (SO, R-4, R-6, R-8, D-4, D-6, D-8), and no curriculum (training each D/R-$n$ task from scratch)—as shown in Tab.~\ref{tab: curriculum}. For the two curricula, we use the best-performing checkpoints from trained 10k iterations for each setting. The curriculum with dense-to-random direction consistently yields the best performance across tasks.
The results indicate that with the progression of the curriculum, the teacher policy demonstrate greater accuracy and efficiency, addressing Q3.

\vspace{-2mm}
\subsection{Generalization to irregular clutters}
\begin{figure}[t!]
    \centering
\includegraphics[width=0.4\textwidth]{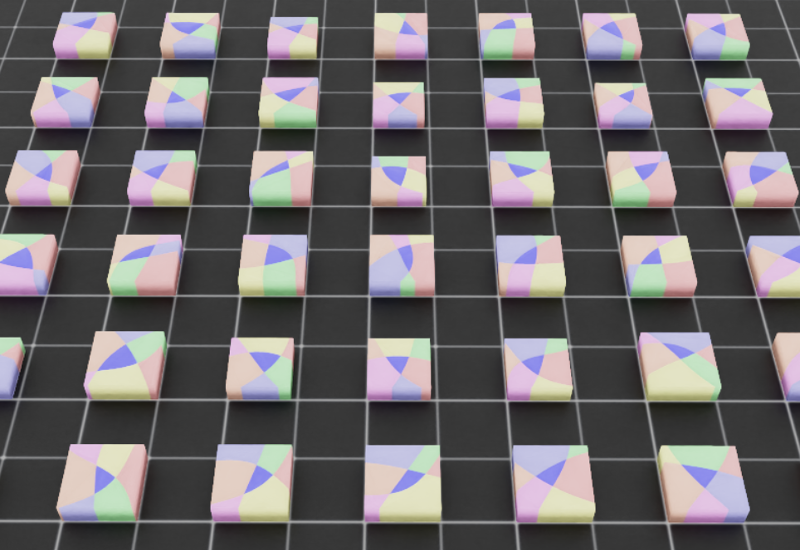}
    \vspace{-1mm}
    \caption{Irregular Clutter. The target object is marked as purple in the center.}
    \label{fig:irregular}
    \vspace{-1mm}
\end{figure}

To test whether our policy generalizes to geometries beyond the cuboid clutter, we evaluate its performance on randomly generated irregular clutters that are tightly packed—a particularly challenging case for SNCs. Each clutter is constructed by dividing a 24×24×8 cm cuboid using three random curves, ensuring three intersections to form seven distinct pieces, shown in Fig.~\ref{fig:irregular}.
We fine-tuned our policy from D-8 expert on 200 such irregular clutters and tested it on 50 unseen instances together with previous baselines. As shown in Table \ref{tab: generalization}, our method achieves good performance, and therefore exhibits geometric generalization, in response to Q4.

\begin{table}[t!]
\centering
\caption{Evaluation on irregular clutters}\label{tab: generalization}
\begin{tabular}{@{}cc|c|c@{}}
\toprule
\multicolumn{2}{c|}{\multirow{1}{*}{Method}} & \multicolumn{1}{c|}{SR(\%)$\uparrow$} & \multicolumn{1}{c}{AS$\downarrow$} \\  \midrule
\multicolumn{1}{c|}{\multirow{3}{*}{State-Based}} & GraspReward-Only          &   26  &   271    \\
\multicolumn{1}{c|}{} & Multi-Stage   &   53  &   245 \\
\multicolumn{1}{c|}{} & Ours (Teacher, D-8 Finetuned)       &   \textbf{81}     &   \textbf{211} \\ \midrule
\multicolumn{1}{c|}{Vision-Based} & Ours (Student, D-8 Finetuned)     &    60    &  231  \\ 
\bottomrule
\end{tabular}
\vspace{-1mm}
\end{table}

\begin{table}[t!]
\centering
\caption{Evaluation of Teacher Policy with Varying Hand Dexterity.}\label{tab: dexterity}
\resizebox{\linewidth}{!}{
\begin{tabular}{@{}c|ccc|c@{}}
\toprule
\multirow{1}{*}{SR(\%)$\uparrow$ } & \multicolumn{1}{c}{D-4} & \multicolumn{1}{c}{D-6} & \multicolumn{1}{c|}{D-8} & \multicolumn{1}{c}{Irregular} \\  \midrule
Low-DoF LEAP Hand                       &   82 (16$\downarrow$)    & 70 (19$\downarrow$) & 71 (16$\downarrow$)  & 24 (57$\downarrow$)\\
Mid-DoF LEAP Hand                       &   91 (7$\downarrow$) & 90 (9$\downarrow$) & 87 (10$\downarrow$)  & 47 (34$\downarrow$)\\
Full-DoF LEAP Hand (Ours)    &   \textbf{98}    & \textbf{99} & \textbf{97}  & \textbf{81} \\ \toprule
\multirow{1}{*}{AS$\downarrow$ } & \multicolumn{1}{c}{D-4} & \multicolumn{1}{c}{D-6} & \multicolumn{1}{c|}{D-8} & \multicolumn{1}{c}{Irregular} \\  \midrule
Low-DoF LEAP Hand                       &    119 (19$\uparrow$)  & 153 (38$\uparrow$) &  166 (28$\uparrow$) &  272 (61$\uparrow$) \\
Mid-DoF LEAP Hand                       &    113 (13$\uparrow$)  & 135 (20$\uparrow$) & 165 (27$\uparrow$)  &  245 (34$\uparrow$) \\
Full-DoF LEAP Hand (Ours)     &   \textbf{100}   & \textbf{115} & \textbf{138}  &  \textbf{211} \\
\bottomrule
\end{tabular}
}
\vspace{-2mm}
\end{table}

\vspace{-2mm}
\subsection{Discussion of Dexterity}\label{sec: dexterity}

In singulation and grasping tasks for SNCs—where limited space restricts finger articulation—the use of high-degree-of-freedom (DoF) hands becomes essential. We evaluate three variants of the LEAP Hand with varying degrees of dexterity:

\begin{itemize}
    \item \textbf{Low-DoF}: The index, middle, and ring fingers are underactuated with fixed abduction–adduction, where each finger’s 3 joints are coupled into 1 DoF. The thumb retains 3 DOFs. (Total: 6 DoF)
    \item \textbf{Mid-DoF}: Index, middle, and ring fingers have independently actuated 3 joints with fixed abduction–adduction. The thumb is fully actuated with 4 DoFs. (total: 13 DoF)
    \item \textbf{Full-DoF}: A fully actuated LEAP Hand with control over all lateral and longitudinal joints of every digit. The thumb is fully actuated with 4 DoFs. (total: 16 DoF)
\end{itemize}

To quantify the impact of dexterity, we benchmarked these 3 embodiments across D-4/6/8 tasks and irregular clutters. The results, summarized in Table~\ref{tab: dexterity}, show significant improvement with increased hand dexterity, in response to Q5. High finger dexterity enables complex contact interactions, which are critical in SNCs where efficient singulation requires delicate finger skills instead of coarse palm movement.

\vspace{-2mm}
\subsection{Real-World Experiments}
\label{sec:real}
\begin{figure}[t!]
    \centering
\includegraphics[width=0.98\linewidth]{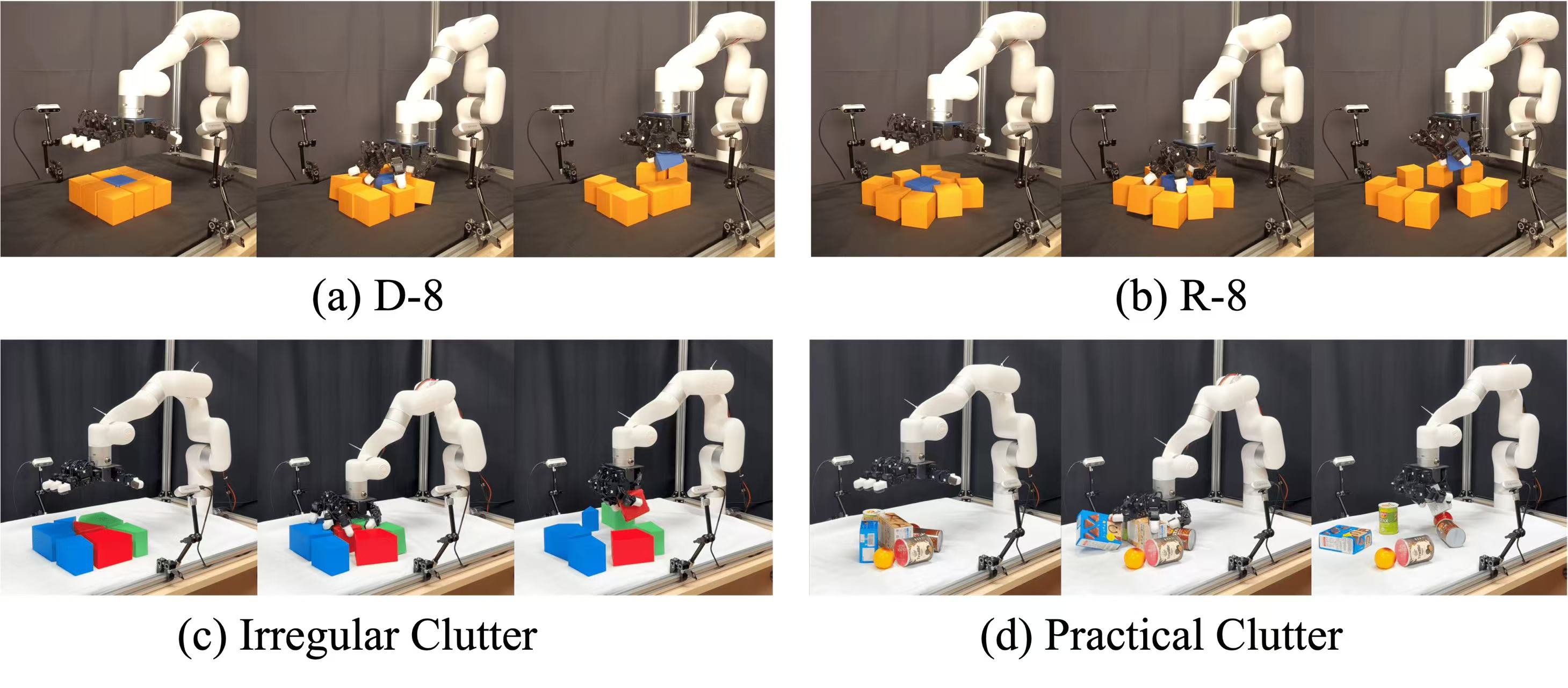}
\vspace{-2mm}
    \caption{Real-world experiments for D-8, R-8,
irregular and practical clutters.}
    \label{fig:grasp_rw}
    \vspace{-1mm}
\end{figure}

\begin{table}[t!]
\centering
\caption{Evaluation of Vision-Based Student Policy in Real-world.}\label{tab:real}
\resizebox{\linewidth}{!}{
\begin{tabular}{c|cccc}
\toprule
Method & D-8  & R-8  & Irregular Clutter & Practical Clutter \\ \midrule
GraspReward-Only  & 3/20 & 2/20 & 0/20              & 17/100     \\ Multi-Stage  & 7/20 & 4/20 & 3/20              & 22/100 \\ Ours & \textbf{15/20} & \textbf{12/20} & \textbf{12/20}             & \textbf{61/100} \\
\bottomrule
\end{tabular}
}
\vspace{-2mm}
\end{table}

We conduct real-world experiments using a uFactory xArm6 robot equipped with a LEAP Hand and two side-view RealSense D435 RGB-D cameras, as shown in Fig.~\ref{fig:grasp_rw}. The synchronized RGB-D data from both cameras are converted into point clouds, transformed to the world frame using calibrated extrinsics, and fused into a single scene point cloud, which is further refined using the Iterative Closest Point (ICP) algorithm. A predefined crop around the clutter center removes background noise, and the fused cloud is downsampled to 1,024 points before being fed into the vision-based policy.

We accomplish singulation and grasping for D-8, R-8, irregular clutter and practical clutter consisting of previously unseen practical objects randomly arranged in the real-world environment and conduct 20 trials for each setting (5 settings for practical clutter).
To further improve sim-to-real transfer, we additionally retrained the teacher policy with domain randomization on hand joint damping, stiffness, and friction (hand friction varies from 0.7 to 1.8 and table friction varies from 0.4 to 1.3 in Isaac Gym). As shown in  Tab.~\ref{tab:real}, the GraspReward-Only baseline struggles in cluttered environments due to the lack of singulation. In contrast, the Multi-Stage method suffers from the sensitivity to sim-to-real discrepancies, often causing failures when singulation generates unseen configurations. Notably, in practical clutter, our policy achieves a 61\% success rate in a zero-shot setting, demonstrating strong generalization across both shape and spatial configurations, even though it was trained solely on a cuboid-shaped curriculum. 

%% file: tex/5-conclusion.tex
Our proposed approach demonstrates that a unified RL framework can effectively integrate object singulation and grasping in SNCs using dexterous hands. By unifying singulation and grasping, our method not only achieves higher grasp success rates and improved efficiency compared to traditional multi-stage approaches but also enables subtle maneuvers directly with finger dexterity. The combination of clutter arrangement curriculum learning and policy distillation further enhances the generalization of the vision-based policy, ensuring successful skill transfer from simulation to the real world, and from seen blocks to unseen objects. Additionally, the introduction of various cluttered grasping environments provides a comprehensive testbed for evaluating performance across various clutter configurations.
Interestingly, a curriculum trained solely on cuboid objects exhibits strong zero-shot generalization to diverse practical clutters, suggesting that spatial relationships—rather than geometric variations alone—are critical to grasping in densely cluttered environments and validating our curriculum design.
 
Future work could extend to more complex clutters, broader object shapes, and improved real-world robustness, while further exploring how spatial diversity in clutter arrangement curricula enhance generalization to unseen clutters.